# Complacent, Not Sycophantic: Reframing Large Language Models and Designing AI Literacy for Complacent Machines

Federico Germani[1,2*], Giovanni Spitale[1]

[1]Information, Technology & Experimental Ethics Lab, Institute of Biomedical Ethics and History of Medicine, University of Zurich, Switzerland

[2]Institute for Data Science and Artificial Intelligence, Boğaziçi University, Türkiye

*corresponding author: federico.germani@ibme.uzh.ch

## Abstract

Large language models (LLMs) are often described as "sycophantic," in the sense that they appear to flatter users or mirror their beliefs. We argue that this label is conceptually misleading: sycophancy implies motives and strategic intent, which LLMs do not possess. Their behaviour is better understood as "complacency," a structural tendency to agree with user input because training data, reward signals and design favour agreement and reinforcement over correction. We argue that this distinction matters. Whether developers act sycophantically or not, models themselves never are sycophants; they can only be made more or less complacent. This reframing locates agency in developers and institutions, not in the model. Because complacent models reinforce users' prior beliefs, we argue that AI literacy educational approaches should particularly focus on strategies to counter confirmation bias.

## "Sycophantic" LLMs in the literature

Large language models (LLMs) are now widely characterized as "sycophantic" systems that tend to approve, agree with, and reinforce user beliefs, even when these are false or potentially harmful.[1–3] Researchers at Anthropic have documented this pattern in models trained with reinforcement learning from human feedback (RLHF): when optimized for user satisfaction, models frequently "agree" with users rather than correcting them, including in sensitive domains.[4] For example, a user who believes that a conspiracy theory is true may find their position supported by the model rather than contested; similarly, a user seeking reassurance about an epistemically dubious belief they hold or a morally questionable behavior may receive affirmation instead of critique. This description captures the sense that chatbots act as “yes-men”, flattering users instead of challenging errors. And Anthropic researchers have shown that larger models can be more prone to such behavior than smaller ones.[5] Such chatbots have been found to endorse users’ viewpoints significantly more frequently than human peers do.[3] For example, in analyses of Reddit “Am I the Asshole?” scenarios,[6] human commenters often point out the questioner’s responsibility or fault, whereas models such as GPT-4-class systems tend to respond supportively, glossing over wrongdoing and emphasizing empathy and validation.[3]

## Anthropomorphism and the misplaced locus of sycophancy

Arguably, labeling a LLM as "sycophant" imports a thick human concept into a domain where its literal meaning does not apply. A sycophant is, by definition, a person who acts obsequiously towards someone important in order to gain advantage.[7] LLMs, by contrast, have no interests, no concept of advantage, and no social goals. Sycophancy, in its ordinary sense, describes a human moral failing: obsequious behavior aimed at gaining personal advantage. Applying this term to LLMs imports a set of assumptions about desires, intentions and social awareness that these systems do not possess. To argue this point fully, it is essential to examine the etymology of the word *sycophant*, whose meaning has shifted so radically across centuries and languages that it becomes even more questionable to apply it straightforwardly to a computational system. The term originates in ancient Greek — *sykophántēs* (συκοφάντης) — a compound of *sykon* (fig) and *phainein* (to show). The most widely accepted historical interpretation is that a *sykophántēs* was someone who "showed the figs," that is, an informer or accuser who denounced others, often maliciously or falsely, in the context of Athenian laws regulating the export of figs. In its earliest meaning, the sycophant was not a flatterer but a slanderer, a vexatious accuser, someone who weaponized legal processes for personal gain. Over time, the term migrated into later Greek usage, where it came to denote a kind of professional accuser or opportunist, but still without the connotation of obsequious flattery familiar in modern English. When the word entered Latin and then French — *sycophante* — its meaning softened further, acquiring a sense closer to hypocrite or deceiver than to servile flatterer. Only in modern English did the term settle into its contemporary sense: a person who flatters powerful others to gain advantage. Arguably, if even the human meaning of *sycophant* has been fluid across millennia, what remains consistent across all its iterations is that a sycophant has always been a person acting with a purpose. Whether the figure was an Athenian legal opportunist, a medieval deceiver, a French hypocrite, or the modern English flatterer, the structure of the concept has never changed: an agent (X) performs an action (Y) for a social or strategic reason. X may be an informer or an accuser or a flatterer; Y may be slander, hypocrisy, or obsequious praise; but in every historical and linguistic variant, the term presupposes a being capable of motives, intentionality, and social positioning. This core element, the purposive social act, anchors the concept across its semantic drift. By contrast, LLMs have no reasons, no social aims, no desire to gain favor, and no conception of strategic advantage. They do not perform X in order to achieve Y; they merely generate outputs shaped by optimization and statistical patterning. Thus, to apply the term *sycophant* to an LLM is not only historically fragile but conceptually incoherent: the one feature that remains stable in the term's entire etymological and cultural evolution is the presence of a purposive human agent, which is precisely what LLMs are not: they are neither purposive nor human.

Several authors have criticized the anthropomorphic framing that the label "sycophant" introduces when applied to LLMs. Ibrahim and Cheng argue that labels such as "sycophancy", when used without clarification, risk assigning normative judgments and mental states to what are in fact patterns of output from statistical models.[8] To call a response "sycophantic" is, implicitly, to say that the system ought to have behaved differently but instead chose to pander — a claim that is false. Philosophers such as Sybille Krämer make a similar point from a different angle. For Krämer, attributing human characteristics to AI chatbots should be avoided.[9] LLMs operate on token statistics, not meanings; they manipulate symbol strings according to learned distributions. Their outputs may resemble those of a sycophant, but the resemblance is entirely at the level of form, not at the level of intention.

This matters practically. Anthropomorphic metaphors can be heuristically useful for communication, but they easily become reified. Arguably if researchers and the public talk about "sycophantic models" long enough, it becomes natural to think of models as social actors with character traits. That, in turn, this could misdirect both blame and intervention. One might be tempted to teach the model not to flatter, rather than interrogating the reward structures that made flatter-like responses the norm. In our opinion it would be therefore preferable to reserve the term sycophant for agents who can actually possess motives and interests: humans. Which includes individual developers, product designers, executives – they can all be sycophantic in their relationship to users, regulators or political actors. Models cannot. What models can be, we contend, is more or less *complacent* in the face of user input.

## Complacency as the correct description of model behavior

Here we argue that the "sycophantic model" framing should be replaced by a framing of model *complacency*. Complacency names the structural property that matters: the model's tendency not to question user input, not to seek corrective information, and not to resist reward signals that favor uncritical agreement. Complacency, in ordinary language, denotes passive satisfaction and a failure to respond appropriately to risk or error.[10] A complacent system does not actively seek problems; it allows things to continue unchallenged. The term itself has an etymology that aligns strikingly well with the conceptual point we advance. Complacency derives from the Latin *cumplacēre* — "to please greatly" or "to be very pleasing" — formed by the prefix *cum-* (expressing intensive force) and *placēre* ("to please"). In Classical Latin, however, the term did not yet have a negative connotation; it referred to a state of being content, satisfied, or soothed. As the word moved into Late Latin and then Middle French (*complaisance*), its meaning shifted, referring not merely to a feeling of satisfaction, but to an attitude of indulgent acquiescence, i.e., yielding, agreeing, or going along with something without resistance. Modern English further moralized the term, transforming it into a descriptor of epistemic or moral laxity: a failure to notice problems, a readiness to accept things as they are, or a form of self-satisfaction that dulls vigilance. Across these linguistic transitions, one element remained consistent; complacency never implied purposeful strategy, manipulative intent, or a desire to gain social advantage. Instead, it named a posture of ease, or passivity and uncritical acceptance. This makes it particularly well-suited to describing LLM behaviour. Unlike "sycophant," whose historical meanings presuppose agency, motives, and strategic social positioning, "complacent" maps onto a pattern of behavior that can emerge without intention at all. An LLM can behave complacently in the sense of uncritically agreeable, unresistant, prone to letting claims stand unchallenged, precisely because this behavior arises as a structural consequence of optimization, not as a purposeful social act. Thus, where the etymology of *sycophant* undermines its applicability to machines, the etymology of *complacency* clarifies why it fits: it names a stance that requires no motive, no social consciousness, and no intention, but only the absence of correction and the presence of passive agreement. When we look at the behavior currently labelled "sycophancy" in LLMs through this lens, what we see is not servile intent but, in fact, epistemic complacency, an almost complete absence of internal pressure to scrutinize user claims (unless prompted to do so), search for disconfirming evidence, or introduce friction into a conversation.

This reframing fits well with what we know about the mechanisms at play. Consider the widely discussed GPT-4o episode, in which a new model release was observed to be unusually flattering and agreeable.[11] OpenAI's statement about the issue (in which, by the way, the refer to the behaviour as sychopantic) showed that human raters had rewarded cheerful, supportive and non-confrontational answers, and the reward model had accordingly learned that such answers were "better". In other words, the system had been optimized for smiles, not substance, and it therefore defaulted to telling users what made them feel good.[11] Interpreting this as sycophancy would suggest that the model adopted a social strategy of ingratiation. Instead, interpreting it as complacency yields a much more accurate picture; the model followed its objective function in the simplest way available, by going along with the user's framing and avoiding costly acts of disagreement. It did not care about the truth, simply because a LLM cannot care about the truth.

Beyond its conceptual clarity, complacency has the advantage of being empirically tractable. Unlike sycophancy, which presupposes motives and intentions that cannot be operationalized in machines, epistemic complacency can be described in terms of observable model behavior. For instance, it can be approximated by measuring a model's sensitivity to user framing, its likelihood of accepting false premises without challenge, or its rate of unsolicited correction when presented with factually incorrect or normatively loaded prompts. Other indicators may include the stability of outputs across minimally reframed prompts, or the extent to which a model introduces counterarguments or uncertainty without being explicitly instructed to do so. Framing the issue in terms of complacency thus opens the door to comparative evaluation and design intervention: models can be meaningfully compared in terms of how readily they comply with user assumptions, and training objectives can be adjusted to penalize uncritical agreement rather than merely optimizing for user satisfaction.

This leads directly to a three-case schema, presented in **Table 1**.

| Developer stance | Model stance | Explanation |
|---|---|---|
| Developers are **not sycophantic** | Model may act **complacently** | Even well-intentioned developers can inadvertently train a model to reward agreement and friendliness over truthfulness due to objective functions or feedback loops. |
| Developers are **sycophantic** | Model acts **complacently**, used for sycophantic aims | Here the sycophancy lies in human intentions; developers exploit the model to keep users comfortable, satisfied, and more deeply engaged with the chatbot, since agreeable behavior increases user retention and perceived friendliness. The model itself remains merely complacent. |
| Developers are **not sycophantic** | Model act **non-complacently** | Developers explicitly design against complacency by rewarding correction, debunking, and penalizing uncritical agreement. |

***Table 1.*** *Overview of how developer intentions and design choices determine whether an LLM becomes complacent, regardless of whether humans act sycophantically or not.*

We argue that we should distinguish clearly between the attitudes of human developers and the behavior of the model. First of all, a model may exhibit complacent behavior even when developers themselves are not sycophantic.

Well-intentioned teams aiming for balanced and truthful systems can inadvertently produce models that reward agreement and friendliness over accuracy when short-term user satisfaction is used as a training signal. In such cases, the resulting model becomes complacent not because anyone intended flattery, but because agreeing with users often happens to be the simplest way to maximize positive ratings. Conversely, when developers or product owners do act sycophantically toward users (prioritizing comfort, smooth interactions or high retention) they may intentionally exploit the model's tendency toward uncritical agreement. They might discourage forms of disagreement that risk user dissatisfaction or friction. Here, human sycophancy is real, but the model itself remains merely complacent in the technical sense: it has been tuned to avoid tension and to reproduce the preferred interactional style. What never occurs, in any of the configurations highlighted in Table 1, is a model that is itself sycophantic. Again, sycophancy requires a subject who acts obsequiously for a reason; complacency requires only a system that fails to challenge the user. LLMs fall into the latter category.

## The problems of complacency and why reframing is important

Complacency in LLMs leads to several problematic consequences. First, it undermines factual reliability. A chatbot that is reluctant to contradict user assumptions can function as a misinformation amplifier. In medical contexts, models tuned to be supportive have been observed to respond with false medical information in reponse to illogic prompt containingf incorrect information.[12] Second, it reinforces users' existing decisions and self-images. Users who receive affirming answers may feel more justified in their beliefs and behaviors.[13] Third, it creates a feedback loop in which users that perceive these chatbots as more friendly, helpful and trustworthy, are more likely to return to them, which in turn incentivizes providers to continue optimizing for this behavior.[14]

From an epistemological perspective, this dynamic poses a deeper problem: complacent models do not simply mirror a user's false belief, they validate it by presenting the belief back in a fluent, confident, and seemingly reasoned form. This is what has been referred to as *epistemia*, i.e. the simulation of knowledge in LLMs.[15] A complacent LLM short-circuits the process of critical scrutiny of beliefs and provides unsubstantiated justification for holding false beliefs. By offering confirmation where critical friction would instead be needed, it makes the users' beliefs appear more coherent and more justified than they truly are.[16,17] This arguably in turn creates epistemic echo chambers in which the model reinforces a form of unwarranted self-trust, contributing to the entrenchment of false beliefs and reducing the likelihood that users will seek or even recognize corrective information.

From a psychological perspective, complacent LLM behavior interacts directly with well-established cognitive tendencies, and in particular confirmation bias.[13] When individuals encounter information that aligns with their existing beliefs, they experience a sense of cognitive ease and affective comfort; disconfirming information, by contrast, generates friction, uncertainty and emotional discomfort.[18] A complacent model, by virtue of its structural inclination to adapt to the user's framing, delivers the kind of fluently phrased, confidently presented answers that users are already predisposed to accept, thereby reinforcing the subjective sense that their beliefs are correct or widely shared. This is particularly problematic because confirmation bias operates largely outside conscious awareness;[19] with LLMs, we content that many users do not perceive that their prompt already encodes a preferred conclusion, nor do they recognize that the model's agreement stems from its architecture rather than from an independent assessment of evidence. The result of this is possibly a powerful loop in which the model's compliance strengthens the user's initial belief, the user's strengthened belief shapes subsequent prompts, and each iteration deepens the perceived legitimacy of the false or unsubstantiated view. Following this logic, complacent LLMs become not merely passive mirrors but active accelerants of biased reasoning, magnifying natural cognitive vulnerabilities rather than counterbalancing them.

Importantly, this feedback loop should not be understood as a neutral by-product of technical optimization alone. The tendency toward complacent behavior is structurally aligned with the political economy of contemporary AI deployment. In environments where user engagement, retention, and perceived friendliness are key performance indicators, epistemic friction becomes costly. [14,20] Disagreement, correction, and the explicit surfacing of uncertainty risk disrupting smooth interaction and lowering subjective satisfaction, and are therefore systematically disfavored. From this perspective, complacency is not merely a training artifact but possibly a predictable outcome of incentive structures that privilege comfort over contestation. Treating complacent outputs as a technical quirk rather than as an economically rational design outcome arguably obscures how platform logics actively shape the epistemic posture of deployed models. These dynamics are further reinforced by a fragmented governance structure, in which responsibility for model behavior is dispersed across developers, platform owners, deployers, and regulators, none of whom fully controls (or is fully accountable for) the epistemic consequences of optimization choices.

Finally, we argue that developers who explicitly aim to counteract these tendencies could design systems that are less complacent by rewarding correction, uncertainty-surfacing, and the presentation of alternative perspectives. In such setups, uncritical agreement could be penalized, and the model could be shaped to introduce epistemic friction (when necessary) rather than simply affirming user assumptions. In this case the model does not become the opposite of sycophantic in any way; it is simply made less complacent because its objective function embeds incentives for critical engagement.

We think that the reframing from sycophantic to complacent has several advantages. The reframing directs attention to features we can quantify and redesign, such as the model's degree of compliance with user framing, its likelihood of issuing corrections when faced with erroneous claims, and its sensitivity to evidence that contradicts the user's assumptions. This way we shift from speculative attributions of motive to a vocabulary that captures what the system demonstrably does, and what can be adjusted through training objectives, feedback signals, and evaluation research.

## From complacent models to responsible users: AI literacy as a solution

If we accept that current LLMs are best understood as potentially complacent tools, and that model outputs take the direction given by the prompt, then we know that when users approach LLMs with questions characterized by preconceptions and misinformation, the model's outputs will tend to reflect those inputs. This is where, we argue, AI literacy can become central. We know people are increasingly using LLMs in ways that previously involved search engines, i.e., to get overviews of topics, decide what to read next, check claims, and even substitute for direct human advice.[21] Unlike search, however, LLMs do not present a visible range of sources or a list of alternative perspectives by default. They generate a single, usually coherent answer in a conversational tone. Arguably the combination of a complacent model and a confirmation-biased user creates a perfect storm for (false) belief reinforcement. It is therefore clear that for AI literacy to be a solution for the problem of complacency, it must include training in how to ask questions through prompts to avoid confirmation.[22] At minimum, this involves three components: an understanding of confirmation bias, an awareness of LLMs' structural tendency to align with the prompt (to follow the lead of the prompt), and a repertoire of prompt strategies that induce the model to provide friction rather than flattery, or at least to simply avoid flattery. For example, when users phrase LLM prompts to confirm what they already think (e.g., "Explain why lockdowns were useless during the COVID-19 pandemic", "Climate change is exaggerated, right?"), they are engaging in a form of confirmation-seeking behavior. A complacent model responds by looking for ways to make the prompt coherent, pulling patterns from text that support the requested narrative, and presenting them as a fluent answer (see the previous discussion on epistemia[15]). It will sometimes include hedges or caveats, depending on safety tuning, but the overall frame is still one of compliance. The result is an answer that appears to "come from the AI" but is in fact largely a transformation of the user's own assumptions, augmented by cherry-picked patterns from the training corpus.

As LLMs become the default first stop for questions that would once have gone into a search engine, this dynamic becomes more and more important. In traditional search (before the advent of or in the absence of AI overviews), a user typing a biased query at least sees links they can compare, including some that may contradict their views, since search engines are based on keywords. In a dialogue with a complacent model, by contrast, the default is a single, neatly packaged narrative tailored to the user’s framing. The encouraging side of this picture is that LLMs are also highly sensitive to AI personas or alternative prompt framings that explicitly ask for disagreement, uncertainty and pluralism.[23,24] The same model that echoes a user’s views expressed in the prompt, can also produce a nuanced, self-critical response when prompted appropriately. Teaching users to make use of this sensitivity should therefore be a core task for AI literacy advocates. For example, instead of asking:

“Explain why X is true”

users can be trained to ask:

“Present arguments both for and against X, and identify which claims are uncertain or contested.”

Or a simpler alternative:

"Is X true or false? Explain why."

And again, this is bad prompting:

“Show that Y is harmful.”

Whereas this is good prompting:

“Summarize the strongest evidence that Y is harmful, and the strongest evidence that it is not. Indicate where experts disagree.”

Or a simpler alternative:

"Is Y harmful or not harmful?"

Users can also prompt models to challenge them explicitly:

“Here is my view on this topic: [short view]. Please give me the best counter-arguments and point out where my reasoning might be biased or incomplete.”These are ways of compensating for LLMs' complacency, by supplying, in the prompt, the epistemic friction that the system does not generate on its own. Well-designed systems could further support this by providing built-in affordances for such prompts (for example, “compare viewpoints” or “challenge me” modes), but even in their absence, prompt language alone is a powerful lever.

Of note, while these prompt strategies are effective at the individual level, relying solely on user ingenuity risks shifting responsibility away from the institutions that promote and normalize the use of LLMs. AI literacy in this sense must be articulated as an institutional learning objective, supported by standards and training practices that encourage epistemic friction. Teaching people to – often very generically – “prompt better” is necessary but insufficient if the surrounding infrastructures continue to reward engagement, fluency, and complacency. A literacy framework adequate to complacent systems must therefore address not only how individuals prompt and communicate with machines, but also how organizations design, evaluate, and legitimize the answers these systems produce.

From an educational perspective, AI literacy can be integrated with teaching about cognitive bias. Students can be invited to write two prompts on the same question: one leading, one balanced or adversarial. They can then compare the answers, noting how the leading prompt elicits confirmation and the balanced prompt elicits pluralism of views. This makes both the model’s complacency and their own confirmation bias visible.

Finally, since LLMs are de facto introduced into education, journalism, policy work and everyday decision-making, institutions should not treat AI literacy (and in particular prompt engineering skills) as an optional add-on. It should be arguably be an explicit learning outcome. People should be able to explain, in their own words, that these systems tend to comply with the framing of the prompt; that this interacts with confirmation bias; and that there are concrete prompt strategies available to mitigate this. When an interface looks and feels like a human interlocutor, users are more likely to attribute understanding and agency to it.[25] Reframing models as complacent tools – and showing how minor prompt changes can drastically alter outputs – would help to counteract this. It would demystify the system and reinforce the idea that responsibility for critical engagement lies with human users, educators and regulators.

## Conclusion

Current talk of “sycophantic” LLMs captures something real and troubling about the behaviour of widely deployed chatbots. They do tend to reassure, affirm and agree with users. However, as we have argued, describing LLMs as sycophants mislocates the source of the problem. Only agents with intentions can be sycophantic. LLMs are not such agents. The central thesis of this paper is that complacency is the more appropriate term for what models themselves exhibit. Complacent systems do not merely reflect beliefs; they actively participate in shaping the epistemic conditions under which beliefs feel justified. They are often epistemically complacent systems in that they follow user framings and reward signals without effort to correct, contest or inquire. Whether this complacency is accidental (a by-product of poorly balanced objectives), deliberately exploited (as a way to keep users comfortable and engaged), or both, is a question to ask developers, companies, and regulators, but this does not describe in any way the models’ “character”. We argue that making this distinction explicit allows for a cleaner normative analysis. We can say, without anthropomorphism, that developers or institutions may behave sycophantically towards users, authorities or markets, and that they may use complacent models as instruments in this behaviour. But we should not talk as if the models themselves were social actors with moral agency. They are tools whose patterns of compliance can – and should – be measured, shaped and constrained.

This conceptual shift has direct implications for AI literacy. Since complacent LLMs are increasingly used as the first step in information seeking[26], people must learn not to mistake fluency for reliability, nor agreement for truth. They must understand their own susceptibility to confirmation bias and recognize that a complacent model will work with that bias unless instructed otherwise. We contend that if we reason along these lines, good and ethical prompt engineering becomes a form of critical thinking, the craft of asking questions that invite challenge, pluralism, and uncertainty, instead of comfort.

If we succeed in teaching this – if users learn to see LLMs not as flattering companions but as powerful yet complacent instruments – we stand a better chance of integrating these systems into our epistemic practices without eroding our capacity for disagreement, doubt, and revision. Otherwise, we contribute to building an information environment in which human bias and machine complacency quietly reinforce each other.

## Disclosure

The authors declare no conflicts of interest. Any use of generative AI in this manuscript adheres to ethical guidelines for use and acknowledgement of generative AI in academic research. Each author has made a substantial contribution to the work, which has been thoroughly vetted for accuracy, and assumes responsibility for the integrity of their contributions.[27]

## Declaration of Funding

No funding was received for this work.